\documentclass{llncs}

\usepackage{graphicx} 
\usepackage{subfigure}
\usepackage{amsmath}
\usepackage{graphicx}
\usepackage{amssymb}
\usepackage{url}
\usepackage[usenames,dvipsnames]{color}
\usepackage{verbatim}

\title{Adaptive Bases for Reinforcement Learning}

\author{Dotan Di Castro \and Shie Mannor}

\titlerunning{Adaptive Bases for Reinforcement Learning}

\toctitle{Adaptive Bases for Reinforcement Learning}

\institute{
Department of Electrical Engineering\\
Technion - Israel Institute of Technology \\ \email{\{dot,shie\}$@$\{tx,ee\}.technion.ac.il}
}

\spnewtheorem{assum}[lemma]{Assumption}{\bfseries}{\itshape}
\spnewtheorem{algo}[lemma]{Algorithm}{\bfseries}{\itshape}
\spnewtheorem{mytheorem}[lemma]{Theorem}{\bfseries}{\itshape}

\begin{document}

\maketitle
\vspace{-0.5cm}

\begin{abstract}
We consider the problem of reinforcement learning using function approximation, where the approximating basis can change dynamically while interacting with the environment. A motivation for such an approach is maximizing the value function fitness to the problem faced. Three errors are considered: approximation square error, Bellman residual, and projected Bellman residual. Algorithms under the actor-critic framework are presented, and shown to converge. The advantage of such an adaptive basis is demonstrated in simulations.
\end{abstract}

\section{Introduction}
Reinforcement Learning (RL) \cite{BerTsi96} is an approach for solving Markov Decision Processes (MDPs), when interacting with an unknown environment. One of the main obstacles in applying RL methods is how to cope with a large state space. In general, the underlying methods are based on dynamic programming, and include adaptive schemes that mimic either value iteration, such as Q-learning, or policy iteration, such as Actor-Critic (AC) methods. While the former attempt to directly learn the optimal value function, the latter are based on quickly learning the value of the currently used policy, followed by a slower policy improvement step. In this paper we focus on AC methods.

There are two major problems when solving MDPs with a large state space. The first is the storage problem, i.e., it is impractical to store the value function and the optimal action explicitly for each state. The second is generalization: some notion of similarity between states is needed since most states are not visited or visited only a few times. Thus, these issues are addressed by the Function Approximation (FA) approach \cite{BerTsi96}, that involves approximating the value function by functional approximators with a smaller number of parameters in comparison to the original number of states. The success of this approach rests mainly on selecting appropriate features, and on a proper choice of the approximation architecture. In a linear approximation architecture, the value of a state is determined by linear combination of the low dimensional feature vector. In the RL context, linear architectures enjoy convergence results and performance guarantees (e.g., \cite{BerTsi96}).

The approximation quality depends on the choice of the basis functions. In this paper we consider the possibility of tuning the basis functions on-line, under the AC framework. As mentioned before, an agent interacting with the environment is composed of two sub-systems. The first is a critic, that estimates the value function for the states encountered. This sub-system acts on a fast time scale. The second is an actor, that based on the critic output, and mainly the \emph{temporal-difference} (TD) signal, improves the agent's policy using gradient methods. The actor operates on a second time scale, slower than the time-scale of the critic. Bhatnagar et al. \cite{BhatnagarEtAl2009Report} proved that such an algorithm with an appropriate relation between the time scales, converges.

We suggest to add a third time scale that is slower than both the critic and the actor, minimizing some error criteria while adapting the critic's basis functions to better fit the problem. Convergence of the value function, policy and the basis is guaranteed in such an architecture, and simulations show that a dramatic improvement can be achieved using basis adaptation.

Using multiple time scales may pose a convergence drawback at first sight. Two approaches may be applied in order to overcome this problem. First, a recent work of Mokkadem and Pelletier \cite{MokPel06}, based on previous research by Polyak \cite{Polyak90} and others, have demonstrated that combining the algorithm iterates with the averaging method of \cite{Polyak90} leads to convergence rate in distribution that is the same as the optimal rate. Second, in multiple time scales the rate between the time steps of the slower and faster time scales should converge to $0$. Thus, time scales which are close, operate on the fast time scale, and satisfy the condition above, are easy to find for any practical needs.

There are several works done in the area of adaptive bases. These works do not address the problem of policy improvement with adaptive bases. We mention here two noticeable works which are similar in spirit to our work. The first work is of Menache et al. \cite{menache2005basis}. Two algorithms were suggested for adaptive bases by the authors: one algorithm is based on gradient methods for \emph{least-squares TD} (LSTD) of Bardtke and Barto \cite{BardtkeBarto1996}, and the other algorithm is based on the cross entropy method. Both algorithms were demonstrated in simulations to achieve better performance than their fixed basis counterparts but no convergence guarantees were supplied.  Yu and Bertsekas \cite{YuBertsekas2009} suggested several algorithms for two main problem classes: policy evaluation and optimal stopping. The former is closer to our work than the latter so we focus on this class. Three target functions were considered in that work: mean TD error, Bellman error, and projected Bellman error.
The main difference between \cite{YuBertsekas2009} and our work (besides the policy improvement) is the following. The algorithmic variants suggested in \cite{YuBertsekas2009} are in the flavor of LSTD and LSPE 
algorithms \cite{Bertsekas07dynamic}, while in our work the algorithms are TD based, thus, in our work no matrix inversion is involved. Also, we demonstrate the effectiveness of the algorithms in the current work.

The paper is organized as follows. In Section \ref{sec:preliminaries} we define some preliminaries and outline the framework. In Section \ref{sec:main_results} we introduce the algorithms suggested for adaptive bases. In Section \ref{sec:analysis} we show the convergence of the algorithms suggested, while in Section \ref{sec:simulations} we demonstrate the algorithms in simulations. In Section \ref{sec:discussion} we discuss the results.

\section{Preliminaries} \label{sec:preliminaries}
In this section, we introduce the framework, review actor-critic algorithms, overview multiple time scales stochastic approximation (MTS-SA), and state a related theorem which will be used later in proving the main results.

\subsection{The Framework} \label{sec-sub:framework}
We consider an agent interacting with an unknown environment that is modeled by a \emph{Markov Decision Process} (MDP) \cite{Puterman1994} in discrete time with a finite state set $X$ and an action set $U$ where $N\triangleq |X|$. Each selected action $u\in U$ of the agent determines a stochastic transition matrix $P_u=[P_u(y|x)]_{x,y\in X}$, where $y$ is the state followed the state $x$.

For each state $x\in X$ the agent receives a corresponding reward $g(x)$ that depend only on the current state\footnote{Generalizing the results presented here to state-action rewards is straight forward.}. The agent maintains a parameterized \emph{policy function} which is a probabilistic function, denoted by $\mu_\theta(u|x)$, mapping an observation $x\in X$ into a probability distribution over the controls $U$. The parameter $\theta\in \bbbr^{K_\theta}$ is a tunable parameter where $\mu_\theta(u|x)$ is a differentiable function w.r.t. $\theta$. We note that for different $\theta$'s, different probability distributions over $\mathcal{U}$ may be associated for each $x\in\mathcal{X}$. We denote by $x_0, u_0, g_0, x_1, u_1, g_1, \ldots$ a state-action-reward trajectory where the subindex specifies time.

Under each policy induced by $\mu_\theta(u|x)$, the environment and the agent induce together a Markovian transition function, denoted by $P_\theta(y|x)$, satisfying $P_\theta(y|x) = \sum_u \mu_\theta(u|x) P_u(y|x)$. The Markovian transition function $P_\theta(y|x)$ induces a
stationary distribution over the state space $X$, denoted by $D(\theta)$. This distribution induces a natural norm, denoted by $\left\Vert \cdot\right\Vert _{D(\theta)}$, which is a weighted norm and is defined by $\left\Vert x\right\Vert_{D(\theta)}^{2}\triangleq x^\top D(\theta)x$. Note that when the parameter $\theta$ changes, the norm changes as well. We denote by $\mathrm{E}_\theta[\cdot]$ the expectation operator w.r.t. the measures $P_\theta(y|x)$ and $D(\theta)$.
There are several performance criteria investigated in the RL literature that differ mainly on their time horizon and the treatment of future rewards \cite{BerTsi96}. In this work we focus on \emph{average reward} criteria defined by
\begin{equation}
  \eta_\theta=\mathrm{E}_\theta[g(x)].
\end{equation}
The agent's goal is to find the parameter $\theta$ that maximizes $\eta_\theta$. Similarly, define the \emph{(differential) value function} as
\begin{equation}
\label{eq:value_func}
J(x)\triangleq\textrm{E}_\theta\left[\left.\sum_{n=0}^{\tau}(g(x_{n})-\eta_\theta)\right|x_{0}=x\right],
\end{equation}
where $\tau\triangleq\min\{k>0|x_{k}=x^{*}\}$ and $x^*$ is some recurrent state for all policies, we assume to exist. Define the \emph{Bellman operator} as $TJ(x)=r-\eta + \mathrm{E}_\theta[J(y)|x]$. Thus, based on \eqref{eq:value_func} it is easy to show the following connection between the average reward to the value function under a given policy \cite{Bertsekas07dynamic}, i.e.,
\begin{equation} \label{eq:BellmanEq}
  J(x) = g(x) - \eta + \mathrm{E}_\theta[J(y)|x] \triangleq TJ(x),
\end{equation}
For later use, we denote by $TJ$ and $J$ the column representations of $J(x)$ and $TJ(x)$ respectively.

We define the Temporal Difference (TD) \cite{BerTsi96,SuttonBarto} of the state $x$ followed by the state $y$ as $d\left(x,y\right) = g(x)-\eta+ J(y)-J(x)$, where for a specific time $n$ we abbreviate $d\left(x_{n},x_{n+1}\right)$ as $d_n$. Based on \eqref{eq:BellmanEq} we can see that
\begin{equation} \label{eq:TD_eq}
\mathrm{E}_\theta[d(x,y)|x]=0,\quad  \textrm{and} \quad \mathrm{E}_\theta[d(x,y)]=0.
\end{equation}
Based on this property, a wide family of algorithms known as TD algorithm exist \cite{BerTsi96}, where common to all these algorithms is solving \eqref{eq:TD_eq} iteratively.

Notational comment: from now on, we omit the dependency on $\theta$ whenever it is clear from the context.

\subsection{Actor-Critic Algorithms}

A well known class of RL approaches is the so called actor-critic (AC) algorithms, where the agent is divided into two components, an actor and a critic. The critic functions as a state value estimator using the so called \emph{TD-learning} algorithm, whereas the actor attempts to select actions based on the TD signal estimated by the critic. These two components solve their own optimization problems separately interacting with each other.

The critic typically uses a function approximator which approximates the value function in a subspace of a reduced dimension $\mathbb{R}^{K_r}$. Define the basis matrix
\begin{equation} \label{eq:phi_mat}
\Phi\triangleq [\phi_k(x_n)]_{1\le n\le N,1\le k\le K_r} \in\mathbb{R}^{N\times K_r},
\end{equation}
where its columns span the subspace $\mathbb{R}^{K_r}$. Thus, the approximation to the value function is $\tilde{J}(x,r) \triangleq \phi \left(x\right)^\top r$, where $r$ is the solution of the following quadratic program $r=\arg\min_{r' \in\mathbb{R}^{K_r}}\left\Vert \Phi r' -J\right\Vert _{D}^{2}$. This solution yields the linear projection operator,
\begin{equation} \label{eq:proj_op_vec}
\Pi= \Phi\left(\Phi ^\top D_\theta\Phi\right)^{-1}\Phi^\top D_\theta
\end{equation}
that satisfies
\begin{equation} \label{eq:projection_operator}
\tilde{J}(r)=\Pi J.
\end{equation}
where $\tilde{J}(r)$ is the vector representation of $\tilde{J}(x,r)$. Abusing notation, we define the (state dependent) projection operator on $J(x)$ as $\tilde J (x) = \Pi J(x)$.

As mentioned above, the actor receives the TD signal from the critic, where based on this signal, the actor tries to select the optimal action. As described in Section \ref{sec-sub:framework}, the actor maintains a policy function $\mu_\theta(u|x)$. In the following, we state a theorem that serves as the foundation for the policy gradient algorithm described later. The theorem relates the gradient w.r.t. $\theta$ of the average reward, $\nabla_\theta \eta_\theta$, to the TD signal, $d(x,y)$. Define the \textit{likelihood ratio derivative} as $\psi_\theta(x,u) \triangleq \nabla_\theta \mu_\theta(u|x)/\mu_\theta(u|x)$.
We omit the dependency of $\psi$ on $x$, $u$, and $\theta$ through that paper. The following assumption states that $\psi$ is bounded.
\begin{assum}
\label{asum:psi_bounded} For all $x\in X$, $u\in U$, and $\theta\in\mathbb{R}^{K_\theta}$,
there exists a positive constant, $B_{\psi}$, such that $\left\Vert \psi\right\Vert_{2},\left\Vert \nabla_\theta\psi\right\Vert_{2}\le B_{\psi}<\infty$.
\end{assum}
Based on this, we present the following lemma that relates the gradient of $\eta$ to the TD signal \cite{BhatnagarEtAl2009Report}.
\begin{lemma}
\label{thrm:grad_eta_based TD}
The gradient of the average reward (w.r.t. to $\theta$) can be expressed by $\nabla_\theta \eta=$\rm{E}$[\psi_\theta(x,u)d(x,y)]$.
\end{lemma}

\subsection{Multiple Time Scales Stochastic Approximation}

Stochastic approximation (SA), and in particular the ODE approach  \cite{KushnerYin2003}, is a widely used method for investigating the asymptotic behavior of stochastic iterates. For example, consider the following stochastic iterate
\begin{equation*}
\varphi_{n+1} = \varphi_{n} + \alpha_{n} G (\varphi_{n},\zeta_{n+1})
\end{equation*}
where $\{\zeta_{n+1}\}$ is some random process and $\{\alpha_n\}$ are step sizes that form a positive series satisfying conditions to be defined later. The key idea of the technique is the following. Suppose that the iterate can be decomposed into a mean function, denoted by $F(\cdot)$, and a noise term (martingale difference noise), denoted by $M_{n+1}$,
\begin{equation} \label{eq:1d_iter}
\varphi_{n+1} = \varphi_{n} + \alpha_{n} G (\varphi_{n}),\zeta_{n+1}) = \varphi_{n} + \alpha_{n} \left( F (\varphi_{n}) + M_{n+1}\right),
\end{equation}
and suppose that the effect of the noise weakens due to repeated averaging. Consider the following ODE which is a continuous version of $\varphi$ and $F(\cdot)$
\begin{equation} \label{eq:1d_ode}
\dot \varphi_t = \left( F (\varphi_t)\right),
\end{equation}
where the dot above a variable stands for a time derivative. Then, a typical result of the ODE method in the SA theory suggests that the asymptotic limit of \eqref{eq:1d_iter} and \eqref{eq:1d_ode} are identical.

The classical theory of SA considers an iterate, which may be in some finite dimensional Euclidean space. Sometimes, we need to deal with several multidimensional iterates, dependent one on the other, and where each iterate operates on different timescale. Surprisingly, this type of SA, called \emph{multiple time scale SA} (MTS-SA), is sometimes easier to analyze, with respect to the same iterates operate on single timescale. The first analysis of two time-scales SA algorithms was given by Borkar in \cite{borkar1997two} and later expanded to MTS by Leslie and Collins in \cite{LeslieCollins2003}. In the following we describe the problem of MTS-SA, state the related ODEs, and finally state the conditions under which MTS-SA iterates converge. We follow the definitions of \cite{LeslieCollins2003}.

Consider $L$ dependent SA iterates as the following
\begin{equation}
\label{eq:MTSSA_iterates}
\varphi_{n+1}^{(i)} = \varphi_{n}^{\left(i\right)} + \alpha_{n}^{\left(i\right)} \left(F^{(i)} \left(\varphi_{n}^{\left(1\right)}, \ldots,\varphi_{n}^{\left(N\right)}\right) + M_{n+1}^{(i)}\right),
\quad
1\le i \le L,
\end{equation}
where $\varphi_{n}^{(i)} \in \mathbb{R}^{d_i}$, and $F^{(i)}: \mathbb{R}^{\otimes_{j=1}^{L} d_j} \rightarrow \mathbb{R}^{d_i}$. The following assumption contains a standard requirement for MTS-SA step size.
\begin{assum} \label{asum:MTS-step-size} (MTS-SA step size assumptions)
\vspace{-2ex}
\begin{enumerate}
\item \label{eq:step_size_conditions} For $1\le n \le L$, we have
    $\sum_{n=0}^{\infty}\alpha_{n}^{\left(i\right)} = \infty, \quad \sum_{n=0}^{\infty}\left(\alpha_{n}^{\left(i\right)}\right)^{2} < \infty,$
\item \label{eq:MTSSA_rates_condition} For $1\le n\le L-1$, we have   $\lim_{n\rightarrow\infty} a_{n}^{\left(i\right)} / {a_{n}^{(i+1)}=0}.$
\end{enumerate}
\end{assum}
We interpret the second requirement in the following way: the higher the index $i$ of an iterate, it operates on higher time scale. This is because that there exists some $n_0$ such that for all $n>n_0$ the step size of the $i$-th iterate is larger uniformly then the step size of the iterates $1\le j \le i-1$. Thus, the $i$-th iterate advances more than any of the iterates $1\le j \le i-1$, or in other words, it operates on faster time scale. The following assumption aggregates the main requirement for the MTS-SA iterates.
\begin{assum} \label{asum:MTS-iterate} (MTS-SA iterate assumptions)
\begin{enumerate}
\item $F^{\left(i\right)}\left(\cdot\right)$ are gloablly Lipschitz continuous,
\item \label{eq:sup_cond} For $1\le i\le L$, we have  $\sup_{n}\left\Vert \varphi_{n}^{\left(i\right)}\right\Vert <\infty$.
\item \label{eq:martingale_convergence} For $1\le i \le L$, $\sum_{k=0}^n a_{k}^{(i)}M_{k+1}^{(i)}$ converges a.s.
\item (The ODEs requirements) Remark: this requirement is defined recursively where requirement (a) below is the initial requirement related to the $L$-th ODE, and requirement (b) below describes the $i$-th ODE system that is recursively based on the $(i+1)$-th ODE system, going from $i=L-1$ to $i=1$. Denote $\varphi^{(i\rightarrow j)}\triangleq (\varphi^{(i)},\ldots,\varphi^{(j)})$.
\begin{enumerate}
\item Define the $L$-th ODE system to be
\begin{equation} \label{eq:ODE_system_L}
\left \{
\begin{array}{ccl}
\dot \varphi^{(1\rightarrow L-1)}_t &=& 0,\\
\dot \varphi^{(L)}_t &=& F^{(L)}(\varphi^{(1)}_t, \ldots, \varphi^{(L)}_t),
\end{array}
\right.
\end{equation}
and suppose the initial condition $ \left.\varphi_t^{(1\rightarrow L-1)} \right|_{t=0} = \varphi_0$. Then, there exists a Lipschitz continuous function $\xi^{(L)}(\varphi_0)$ such that the ODE system \eqref{eq:ODE_system_L} converges to the point $(\varphi_0,\xi^{(L)}(\varphi_0))$.

\item Define the $i$-th ODE system, $i=L-1, \ldots, 1$, to be
\begin{equation} \label{eq:ODE_system_i}
\left \{
\begin{array}{ccl}
\dot \varphi^{(1\rightarrow i-1)}_t &=& 0,\\
\dot \varphi^{(i)}_t &=& F^{(i)}(\varphi^{(1)}, \ldots,\varphi^{(i-1)}, \varphi^{(i)}, \xi^{(i+1 )}(\varphi_0,\varphi^{(i)})),
\end{array}
\right.
\end{equation}
where $\xi^{(i+1)}(\cdot,\cdot)$ is determined by the $(i+1)$-th ODE system, and suppose the initial condition $ \left.\varphi^{(1\rightarrow i-1)}_t \right|_{t=0}= \varphi_0 $. Then, there exists a Lipschitz continuous function $\xi^{(i)}(\varphi_0 )$ such that the ODE system \eqref{eq:ODE_system_i} converges to the point $(\varphi_0,\xi^{(i)})$.

\end{enumerate}
\end{enumerate}
\end{assum}
The first two requirements are common conditions for SA iterates to converge. The third requirement ensures the noise term asymptotically vanishes. The fourth requirement ensures (using a recursive definition) that for each time scale $i$, where the slower time scales $1,\ldots,i-1$ are static and where for the faster time scales $i+1,\ldots,L$ there exists a function $\xi^{(j+1\rightarrow L)} (\cdot)$ (which is the solution of the  $i+1$ ODE system), there exists a Lipschitz convergent function. Based on these requirements, we cite the following theorem due to Leslie and Collins \cite{LeslieCollins2003}.
\begin{mytheorem}
\label{thrm:MTS}Consider the iterate \eqref{eq:MTSSA_iterates}
and suppose Assumption \ref{asum:MTS-step-size} and \ref{asum:MTS-iterate} hold. Then, the asymptotic behavior of the iterates \eqref{eq:MTSSA_iterates} converge to the invariant set of the dynamic system
\begin{equation}
\dot \varphi_t^{(1)} = F^{(1)}\left(\varphi_t^{(1)},\xi^{(2)} \left(\varphi_t^{(1)}\right)\right),
\end{equation}
where $\xi^{(2)}(\cdot)$ is determined by requirement 4 of Assumption \ref{asum:MTS-iterate}.
\end{mytheorem}

\section{Main Results} \label{sec:main_results}
In this section we present the main theoretical results of the work. We start by introducing adaptive bases and show the algorithms that are derived from choosing different approximating schemes.

\subsection{Adaptive Bases}

The motivation for adaptive bases is the following. Consider an agent that chooses a basis for the critic in order to approximate the value function. The basis which one chooses with no prior knowledge might not be suitable for the problem at hand. A poor subspace where the actual value function is poorly supported may be chosen. Thus, one might prefer to choose a parameterized basis that has additional flexibility by changing a small set of parameters.

We propose to consider a basis that is linear in some of the parameters but has several other parameters that allow greater flexibility. In other words, we consider bases that are linear with respect to some of the terms (related to the fast time scale), and nonlinear with respect to the rest (related to the slow time scale). The idea is that most probably one does not lose from such an approach in general if it fails, but in many cases it is possible to obtain better fitness and thus a better performance, due to this additional flexibility. Mathematically,
\begin{equation}
\label{eq:J_SLFA}
\tilde{J}(x,r,s)=\phi\left(x,s\right)^\top r,\quad s\in\mathbb{R}^{K_s},
\end{equation}
where $r$ is a linear parameter related to the fast time scale, and $s$ is the non-linear parameter related to the slow time scale. In the view of \eqref{eq:phi_mat}, we note  that from now on the matrix $\Phi$ depends on $s$, i.e., $\Phi\equiv \Phi_s$, and in matrix form we have $\tilde J=\Phi_s r$, but for ease of exposition we drop the dependency on $s$. The following assumption is needed for proving later results.
\begin{assum} \label{asum:phi_Lipschitz_bounded}
The columns of the the matrix $\Phi$ are linearly independent, $K_r<N$, and $\Phi r\ne e$, where $e$ is a vector of $1$'s. Moreover, the functions $\phi\left(x,s\right)$ and $\partial \phi\left(x,s\right)/\partial s_i$ for $1\le i \le K_s$ are Liphschitz in $s$ with a coefficient $L_\phi$, and bounded with coefficient $B_\phi$.
\end{assum}
Notation comment: for ease of exposition, we drop the dependency on $x_n$, e.g., $\phi_n\equiv \phi(x_n,s_n)$, $g_n\equiv g(x_n)$. Denote $\phi\triangleq\phi(x,s)$, $\phi'\triangleq\phi(y,s)$ (where as in Section \ref{sec-sub:framework}, $y$ is the state followed the state $x$), $\phi_n'\triangleq \phi(x_{n+1},s_n)$, $d_n \triangleq d(x_n,x_{n+1})$, and $d\triangleq d(x,y)$. Thus, $d = g - \eta + \phi'^\top r - \phi^\top r$ and $d_n = g_n - \eta_n + \phi_{n}'^\top r_n - \phi_n^\top r_n$.

\subsection{Minimum Square Error and TD}
Assume a basis parameterized as in \eqref{eq:J_SLFA}. The \emph{minimum square error} (MSE) is defined as
\begin{equation*}
\textrm{MSE}=
\frac{1}{2}\mathrm{E}\left[\left(\tilde{J}(x)-J(x)\right)^2\right].
\end{equation*}
The gradient with respect to $r$ is
\begin{equation} \label{eq:td_nabla_r}
\nabla_r \textrm{MSE}=
\frac{1}{2}\mathrm{E}\left[\left(\tilde{J}(x)-J(x)\right)\phi\right]
\approx
\textrm{E}\left[ d \phi \right],
\end{equation}
where in the approximation we use the bootstrapping method (see \cite{SuttonBarto} for a disussion) in order to get the well known TD algorithm (i.e., substituting $J\approx T\tilde{J}$). On top of the above TD algorithm, we take a derivative with respect to $s_i$, $i=1,\ldots,K_s$, yielding
\begin{equation} \label{eq:td_partial_s}
\frac{\partial \textrm{MSE}}{\partial s_{i}}
= \textrm{E} \left[ \left(\tilde{J}(x)-J(x)\right) \frac{\partial \tilde{J}(x)}{\partial s_{i}} \right]
\approx
\mathrm{E}\left[d\frac{\partial \phi^\top}{\partial s_{i}}  r\right],
\end{equation}
where again we use the bootstrapping method. Note that this equation gives the non-linear TD procedure for the basis parameters. We use SA in order to solve the stochastic equations \eqref{eq:td_nabla_r} and \eqref{eq:td_partial_s}, which together with Theorem \ref{thrm:grad_eta_based TD} is the basis for the following algorithm. For technical reasons, we add an requirement that the iterates for $\theta$  and $s$ are bounded, which practically is not constraining (see \cite{KushnerYin2003} for discussion on constrained SA).
\begin{algo} \label{algo:TD}Adaptive basis TD (ABTD).
\begin{eqnarray}
\label{eq:TD_iterate_eta}
\eta_{n+1} &=& \eta_n + \alpha_n^{(3)}  \left(g_n-\eta_n  \right),\\
\label{eq:TD_iterate_r}
r_{n+1} &=& r_n + \alpha_n^{(3)} d_n \phi_n, \\
\label{eq:TD_iterate_theta}
\theta_{n+1} &=& H_P^{(\theta)}\left[\theta_n +  \alpha_n^{(2)} \psi_{n} d_n\right],\\
\label{eq:TD_iterate_s}
s_{i,n+1} &=& H_P^{(s)}\left[s_{i,n} + \alpha_n^{(1)} d_n\frac{\partial \phi_n^\top}{\partial s_{i}}  r_n\right],\quad i=1,\ldots,K_s,
\end{eqnarray}
where $H_P^{(\theta)}$ and $H_P^{(s)}$ are projection operators into a non-empty open constraints set whenever $\theta_n\notin H_p$ and $s\notin H_s$, respectively, and the step size series $\{\alpha_n^{(i)}\}$ for $i=1,2,3$ satisfy Assumption \ref{asum:MTS-step-size}.
\end{algo}
We note that this algorithm is an AC algorithm with three time scales: the usual two time scales, i.e., choosing $\{\alpha_n^{(1)}\}_{n=1}^{\infty}\equiv 0$ yields Algorithm 1 of \cite{BhatnagarEtAl2009Report}, and the third iterates is added for the basis adaptation, which is the slowest.

\subsection{Minimum Square Bellman Error}
The Minimum Square Bellman Error (MSBE) is defined as
\begin{equation*}
\textrm{MSBE}=\frac{1}{2}\textrm{E} \left [\left(T\tilde J(x) - \tilde J(x)\right)^2 \right].
\end{equation*}
The gradient with respect to $r$ is
\begin{equation*} \label{eq:MSBE_nabla_r}
\nabla_r \textrm{MSBE}=\textrm{E} \left[d\left(\phi'-\phi \right)\right],
\end{equation*}
where the derivative with respect to $s_i$, $i=1,\ldots,K_s$, is
\begin{equation*}
\frac{\partial \textrm{MSBE}}{\partial s_i} = \textrm{E} \left[d\left(\frac{\partial \phi'^\top}{\partial s_i} - \frac{\partial \phi^\top}{\partial s_i} \right)r \right].
\end{equation*}
Based on this we have the following SA algorithm, that is similar to Algorithm \ref{algo:TD} except for the iterates for $r_n$ and $s_n$.
\begin{algo} \label{algo:MSBE} - Adaptive Basis for Bellman Error (ABBE). Consider the iterates for $\eta$ and $\theta$ in Algorithm \ref{algo:TD}. The iterates for $r$ and $s_i$ are
\begin{equation*}
\begin{split}
r_{n+1} &= r_n - \alpha_n^{(3)} d_n \left( \phi_n'- \phi_n\right), \\
s_{i,n+1} &= \!H_P^{(s)}\left[s_{i,n} - \alpha_n^{(1)} d_n\left( \frac{\partial \phi_n'}{\partial s_{i}}   -  \frac{\partial \phi_n}{\partial s_{i}}  \right)^\top r_n\right],\quad i=1,\ldots,K_s.
\end{split}
\end{equation*}
\end{algo}

\subsection{Minimum Square Projected Bellman Error}
The Minimum Square Projected Bellman Error (MSPBE) is defined as
\begin{equation*}
\textrm{MSPBE}=\textrm{E} \left[ \left(\Pi T\tilde J(x) - \tilde J(x) \right)^2 \right]=\textrm{E} \left[d\phi\right]'\left(\textrm{E} \left[\phi\phi'\right]\right)^{-1} \textrm{E}\left[d\phi\right],
\end{equation*}
where the projection operator is defined in \eqref{eq:proj_op_vec} and where the second equality was proved by Sutton et al. \cite{SuttonEtAl2009ICML}, Section 4. We note that the projection operator is independent of $r$ but depend on the basis parameter $s$. Define $w=\left(\textrm{E}\left[\phi\phi'\right]\right)^{-1}\textrm{E}\left[d\phi\right]$.
Thus, $w$ is the solution to the equation
$\left( \textrm{E} \left[\phi\phi' \right] \right) w = \textrm{E}\left[d\phi\right]$, which yields $\textrm{MSPBE}=w'\textrm{E}\left[d\phi\right]$. Define similar to \cite{BerTsi96} section 6.3.3 $Ar +b \triangleq \textrm{E}\left[d\phi\right]$, where $A=\textrm{E}[\phi(\phi'-\phi)^\top]$ and $b=\textrm{E}[\phi(g-\eta) ]$. Define $A^{\left(i\right)}$ to be the $i$-th column of $A$. For later use, we give here the gradient of $w$ with respect to $r$ and $s$ in implicit form
\begin{equation*}
\begin{split}
\left(\textrm{E} \left[\phi\phi^\top\right] \right)\frac{\partial}{\partial r_{i}}w  &= A^{\left(i\right)},\\
\textrm{E}\left[\phi\phi^\top\right]
\frac{\partial}{\partial s_{i}}w +
\frac{\partial}{\partial s_{i}}\textrm{E}\left[\phi\phi^\top\right]w  &=  \frac{\partial A}{\partial s_{i}}r+\frac{\partial b}{\partial s_{i}}.
\end{split}
\end{equation*}
Denote by $A_{n}$, $A_{i,n}^{s}$,$b_{i,n}^{s}$,$w_{n}$,
$w_{i,n}^{r}$, and $w_{i,n}^{s}$ the estimators at time $n$ of $A$, $\partial A /\partial s_{i}$,
$\partial b /\partial s_{i}$, $w$, $\partial w /\partial r_{i}$, and $\partial w/\partial s_{i}$, respectively.  Define $A_{n}^{\left(i\right)}$ to be the $i$-th column of $A_{n}$. Thus, the SA iterations for these estimators are
\begin{eqnarray*}
\label{eq:estimator_A}
A_{n+1} & = & A_{n}+\alpha_{n}^{(4)} \left(\phi_{n} \left(\phi_{n}-\phi_{n+1}\right)^\top -A_{n}\right),\\
\label{eq:estimator_A_theta}
A_{i,n+1}^{s} & = & A_{i,n}^{s}+\alpha_{n}^{(4)}\left(\frac{\partial \phi_{n}}{\partial s_{i}}\left(\phi_{n}-\phi_{n+1}\right)^\top \right.
\left . + \phi_{n}\frac{\partial}{\partial s_{i}}\left(\phi_{n}-\phi_{n+1}\right)^\top -A_{i,n}^{s}\right),\\
\label{eq:estimator_b_theta}
b_{i,n+1}^{s} & = & b_{i,n}^{s}+\alpha_{n}^{(4)}\left(g\frac{\partial \phi_{n}}{\partial s_{i}}-b_{i,n}^{s}\right),\\
\label{eq:estimator_w}
w_{n+1} & = & w_{n}+\alpha_{n}^{(4)}\left(\phi_{n}d_{n}-\phi_{n}\phi_{n}^\top w_{n}\right),\\
\label{eq:estimator_w_r}
w_{i,n+1}^{r} & = & w_{i,n}^{r} + \alpha_{n}^{(4)} \left(A_{n}^{\left(i\right)}-\phi_{n}\phi_{n}^\top w_{i,n}^{r}\right), \\
\label{eq:estimator_w_theta}
w_{i,n+1}^{s} & = & w_{i,n}^{s}+\alpha_{n}^{(4)} \left(A_{i,n}^{s}r_{n}+b_{i,n}^{s}\right.
 \left.-\left(\frac{\partial}{\partial s_{i}}\left(\phi_{n}\phi_{n}^\top \right)\right)w_{n}-\phi_{n}\phi_{n}^\top w_{i,n}^{s}\right).
\end{eqnarray*}
where $\left\{ \alpha_{n}^{(4)}\right\}$ satisfies Assumption \ref{asum:MTS-step-size}. Next, we compute the gradient of the objective function $\textrm{MSPBE}$ with respect to $r$ and $s$ and suggest a gradient descent algorithm to find the optimal value. Thus,
\begin{eqnarray*}
\frac{\partial\textrm{MSPBE}}{\partial r_{i}}
& = & \textrm{E}\left[d\phi\right]^\top \frac{\partial}{\partial r_{i}}w^\top +w^\top \frac{\partial}{\partial r_{i}}\textrm{E}\left[d\phi\right],\\
\frac{\partial\textrm{MSPBE}}{\partial s_{i}}
& = &
\frac{\partial w^\top}{\partial s_{i}} \textrm{E}\left[d\phi\right]+w^\top \frac{\partial \textrm{E}\left[d\phi\right]}{\partial s_{i}}.\end{eqnarray*}
The following algorithm gives the SA iterates for $r$ and $s$, where the iterates for $\eta$ and $\theta$ are the same as in Algorithms \ref{algo:TD} and \ref{algo:MSBE} and therefore omitted. This algorithm has four time scales. The fastest time scale, related to the step sizes $\{\alpha_n^{(4)}\}$, is the estimators time scale, i.e., the estimators for $A$, $\partial A /\partial s_{i}$, $\partial b /\partial s_{i}$, $w$, $\partial w /\partial r_{i}$, and $\partial w/\partial s_{i}$. The linear parameters of the critic, i.e., $r$ and $\eta$, related to the step sizes $\{\alpha_n^{(3)}\}$, estimated on the second fastest time scale. The actor parameter $\theta$, related to the step sizes $\{\alpha_n^{(2)}\}$, is estimated on the second slowest time scale. Finally, the     critic non-linear parameter $s$, related to the step sizes $\{\alpha_n^{(1)}\}$, is estimated on the slowest time scale. We note that a version where the two fastest times scales operate on a joint single fastest time scale is possible, but results additional technical difficulties in the convergence proof.
\begin{algo}\label{algo:ABPBE} - Adaptive Basis for PBE (ABPBE). Consider the iterates for $\eta$ and $\theta$ in Algorithm \ref{algo:TD}. The iterates for $r$ and $s$ are
\begin{eqnarray*}
r_{i,n+1} & = & r_{i,n} - \alpha_{n}^{(3)}\left(d_{n}\phi_{n}^\top w_{i,n}^{r}+w_{n}^\top A_{n}^{\left(i\right)}r_{i,n} r_{n}\right),\\
s_{i,n+1} & = & s_{i,n} - \alpha_{n}^{(1)}\left(d_{n}\phi_{n}^\top w_{i,n}^{s}+\left(A_{i,n}^{s}r_{n}+b_{i,n}^{s}\right)^\top w_{n}\right),\quad i=1,\ldots,K_s.
\end{eqnarray*}
\end{algo}

\section{Analysis} \label{sec:analysis}
In this section we prove the convergence of the previous section Algorithm \ref{algo:TD} and \ref{algo:MSBE}. We omit the convergence proof of Algorithm \ref{algo:ABPBE} that is similar to the convergence proof of Algorithm \ref{algo:MSBE}.

\vspace{-0.5cm}

\subsection{Convergence of ABTD}
We begin by stating a theorem regarding the ABTD convergence. Due to space limitations, we give only a proof sketch based on the convergence proof of Theorem 2 of Bhatnagar et al. \cite{BhatnagarEtAl2009Report}. The self-contained proof under more general conditions is left to the long version of this work.
\begin{mytheorem} \label{thrm:TD_convergence}
  Consider Algorithm \ref{algo:TD} and suppose Assumption \ref{asum:psi_bounded}, \ref{asum:MTS-step-size}, and \ref{asum:phi_Lipschitz_bounded}, hold. Then, the iterates \eqref{eq:TD_iterate_eta}-\eqref{eq:TD_iterate_s} of Algorithm \ref{algo:TD} converge w.p. 1 to a point that locally maximizes $\eta$ and solves the equation $\rm{E}$$[d\nabla_s  \phi^\top  r]=0$.
\end{mytheorem}
\begin{proof} (Sketch)
There are three time-scales in \eqref{eq:TD_iterate_eta}-\eqref{eq:TD_iterate_s}, therefore, we wish to use Theorem \ref{thrm:MTS}, i.e., we need to prove that the requirements of Assumption \ref{asum:MTS-iterate} are valid w.r.t. to all iterations, i.e., $\eta_n$, $r_n$, $\theta_n$, and $s_n$.

\noindent \textbf{Requirement 1-4 w.r.t. iterates $\eta_n$, $r_n$, $\theta_n$.} Bhatnagar et al. proved in \cite{BhatnagarEtAl2009Report} that \eqref{eq:TD_iterate_eta}-\eqref{eq:TD_iterate_theta} converge for a specific $s$. Assumption \ref{asum:phi_Lipschitz_bounded} implies that the requirements 1-4 of Assumption \ref{asum:MTS-iterate} are valid regarding the iterates of $\eta_n$, $r_n$ and $\theta_n$ uniformly for all $s\in \bbbr^{K_s}$. Therefore, it sufficient to prove that on top of \eqref{eq:TD_iterate_eta}-\eqref{eq:TD_iterate_theta} also iterate \eqref{eq:TD_iterate_s} converges, i.e., that requirements 1-4 of Assumption \ref{asum:MTS-iterate} are valid w.r.t. $s_n$.

\noindent \textbf{Requirement 1 w.r.t. iterate $s_n$.} Define the $\sigma$-algebra $\mathcal{F}_n \triangleq
\sigma(\eta_k, r_k, \theta_k, s_k: k\le n)$, and define
$F_n^{(\eta)} \triangleq \mathrm{E} [g_n-\eta_n|\mathcal{F}_n]$, $F_n^{(r)} \triangleq \mathrm{E} [d_n \phi_n|\mathcal{F}_n]$, $F_n^{(\theta)} \triangleq H_P^{(\theta)}\mathrm{E} [\psi_{n} d_n|\mathcal{F}_n]$, $F_n^{(s_i)} \triangleq H_P^{(s)}\mathrm{E} [d_n\frac{\partial \phi_n^\top}{\partial s_{i}}  r_n|\mathcal{F}_n]$, and $M_{n+1}^{(s_i)} \triangleq H_P^{(s)}[(d_n\frac{\partial \phi_n^\top}{\partial s_{i}}  r_n)-F_n^{(s_i)}]$. Thus, \eqref{eq:TD_iterate_s} can be expressed as
\vspace{-1ex}
\begin{equation}
\label{eq:TD_iterate_s_m}
s_{i,n+1} = s_{i,n} + \alpha_n^{(1)} \left(F_n^{(s_i)} + M_{n+1}^{(s_i)} \right).
\end{equation}
Trivially, using Assumption \ref{asum:phi_Lipschitz_bounded}, $F_n^{(r)}$, $F_n^{(\theta)}$, and $F_n^{(s)}$ are Liphschitz, with respect to $s$, with coefficients $B_\phi^2$, $L_\phi$, and $L_\phi$, respectively. Also, $F_n^{(s_i)}$ is Liphschitz with respect to $\eta$, $r$, and $\theta$ with coefficients $1$, $B_\phi$, and $1$, respectively. Thus, requirement 1 of Assumption \ref{asum:MTS-iterate} is valid.

\noindent \textbf{Requirements 2 and 3 w.r.t. iterate $s_n$.} By construction, the iterate $s_n$ is bounded. Requirement 3 of Assumption \ref{asum:MTS-iterate} is valid  using the boundedness of the martingale difference noise $M_{n+1}^{(s_i)}$ that implies, using the martingale convergence theorem \cite{BerTsi96}, that the martingale $\sum_n \alpha_n^{(3)} M_{n+1}^{(s_i)}$ converges.

\noindent \textbf{Requirement 4 w.r.t. iterate $s_n$.} Using the result of Bhatnagar et al. \cite{BhatnagarEtAl2009Report}, the fast time scales converge w.r.t. the slow time scale. Thus, Requirement 4 is valid based on the fact that the iterates \eqref{eq:TD_iterate_eta}-\eqref{eq:TD_iterate_theta} converge.\qed
\end{proof}

\subsection{Convergence of Adaptive Basis for Bellman Error}
We begin by stating the theorem and then we prove it.
\begin{mytheorem} \label{thrm:ETDsquare}
Consider Algorithm \ref{algo:MSBE} and suppose that Assumption \ref{asum:psi_bounded}, \ref{asum:MTS-step-size}, and \ref{asum:phi_Lipschitz_bounded}, hold. Then, Algorithm \ref{algo:MSBE} converge w.p. 1 to a point that locally maximizes $\eta$ and locally minimizes $\rm{E}$$[d^2]$.
\end{mytheorem}
\begin{proof} (Sketch) To use Theorem \ref{thrm:MTS} we need to check that Assumption \ref{asum:MTS-iterate} is valid. Define the $\sigma$-algebra $\mathcal{F}_n \triangleq
\sigma(\eta_k, r_k, \theta_k, s_k: k\le n)$, and define
$F_n^{(\eta)} \triangleq \mathrm{E} [g_n-\eta_n|\mathcal{F}_n]$, $M_{n+1}^{(\eta)} \triangleq (g_n-\eta_n)-F_n^{(\eta)}$, $F_n^{(r)} \triangleq - \mathrm{E} [d_n (\phi_{n+1}-\phi_{n})|\mathcal{F}_n]$, $M_{n+1}^{(r)} \triangleq -(d_n (\phi_{n+1}-\phi_{n}) ) - F_n^{(r)}$, $F_n^{(\theta)} \triangleq \mathrm{E} [\psi_{n} d_n|\mathcal{F}_n]$, $M_{n+1}^{(\theta)} \triangleq (\psi_{n} d_n)-F_n^{(\theta)}$, $F_n^{(s_i)} \triangleq -\mathrm{E} [d_n(\frac{\partial \phi_{n+1}^\top}{\partial s_{i}}  r_n - \frac{\partial \phi_n^\top}{\partial s_{i}}  r_n)|\mathcal{F}_n]$, and $M_{n+1}^{(s_i)} \triangleq -(d_n (\frac{\partial \phi_{n+1}^\top}{\partial s_{i}}  r_n) - \frac{\partial \phi_n^\top}{\partial s_{i}}  r_n) )-F_n^{(s_i)}$.

On the fast time scale (which is related to $a_n^{(3)}$), as in Theorem \ref{thrm:TD_convergence}, $\eta_n$ converges to $\mathrm{E}[g(x)]$. On the same time scale we need to show that  the iterate for $r_n$ converges. Using the above definitions, we can write the iteration $r_n$ as
\begin{equation}\label{eq:iter_r2}
  r_{n+1} = r_n + \alpha_n^{(3)} \left(F_n^{(r)}+M_{n+1}^{(r)} \right).
\end{equation}
We use Theorem 2.2 of Borkar and Meyn \cite{BorkarMeyn2000ode} to achieve this. Briefly, this theorem states that given an iteration as \eqref{eq:iter_r2}, this iteration is bounded w.p.1 if
\begin{description}
  \item[(A1)] The process $F_n^{(r)}$ is Lipschitz, the function $F_\infty(\sigma) \triangleq\lim_{\sigma\rightarrow \infty} F^{(r)}(\sigma r)/r$ is Lipschitz, and $F_\infty(\sigma)$ is asymptotically stable in the origin.
  \item[(A2)] The sequence $M_{n+1}^{(r)}$ is a martingale difference noise and for some $C_0$
\begin{equation*}
 \textrm{E} \left[(M_{n+1}^{(r)})^2 |\mathcal {F}_n \right] \le C_0 (1+\|r_n\|^2).
\end{equation*}
\end{description}
Trivially, the function $F_n^{(r)}$ is Lipschitz continuous, and we have
\[\lim_{\sigma\rightarrow \infty} F^{(r)}(\sigma r)/r = -\mathrm{E} \left[(\phi'-\phi)(\phi'-\phi)^\top| \right] r.\] Thus, it is easy to show, using Assumption \ref{asum:phi_Lipschitz_bounded}, that the ODE $\dot r =F_\infty^{(r)}$ has a unique global asymptotically stable point at the origin and (A1) is valid. For (A2) we have
\begin{equation*}
\begin{split}
\mathrm{E} & \left[\left. \left\|M(n+1)^{(r)}\right\|^2 \right| \mathcal{F}_n \right] \le \mathrm{E} \left[\left. \left\|d_n \left( \phi'_{n}-\phi_n \right)\right\|^2 \right| \mathcal{F}_n \right] \\
& \le 2 \left( B_g + B_\eta +4 B_\phi^2 r_n\right)^2 \triangleq K''(1+\left\|r_n \right\|^2),
\end{split}
\end{equation*}
where the first inequality results from the inequality $\mathrm{E}[(x-\mathrm{E}[x])^2]\le \mathrm{E}[x^2]$, and the second inequality results from the uniform boundedness of the involved variables. We note that the related ODE for this iteration is given by $\dot r = F^{(r)}$, and the related Lyapunov function is given by $\textrm{E} [d^2]$. Next, we need show that under the convergence of the fast time scales for $\eta_n$ and $r_n$, the slower iterate for $\theta$ converges. The proof of this is identical to that of Theorem 2 of \cite{BhatnagarEtAl2009Report} and is therefore omitted. We are left with proving that if the fast timescales converge, i.e., the iterates $\eta_n$, $r_n$, and $\theta_n$, then the iterate $s_n^{(i)}$ converge as well. The proof follows similar lines as of the proof for $s_n^{(i)}$ in the proof of Theorem \ref{thrm:TD_convergence}, whereas here the iterate $s_n$ converge to the stable point of the ODE $\dot s=\nabla_s \textrm{E}[d(x,y)^2]$. \qed
\end{proof}

\vspace{-1cm}

\section{Simulations} \label{sec:simulations}
In this section we report empirical results applying the algorithms on two types of problems: \textsc{Garnet} problems \cite{Archibald_etal1995} and the mountain car problem.

\vspace{-0.4cm}

\subsection{\textsc{Garnet} problems} 
\begin{figure}[ht]
\begin{center}
\centerline{\includegraphics[width=0.35\columnwidth]{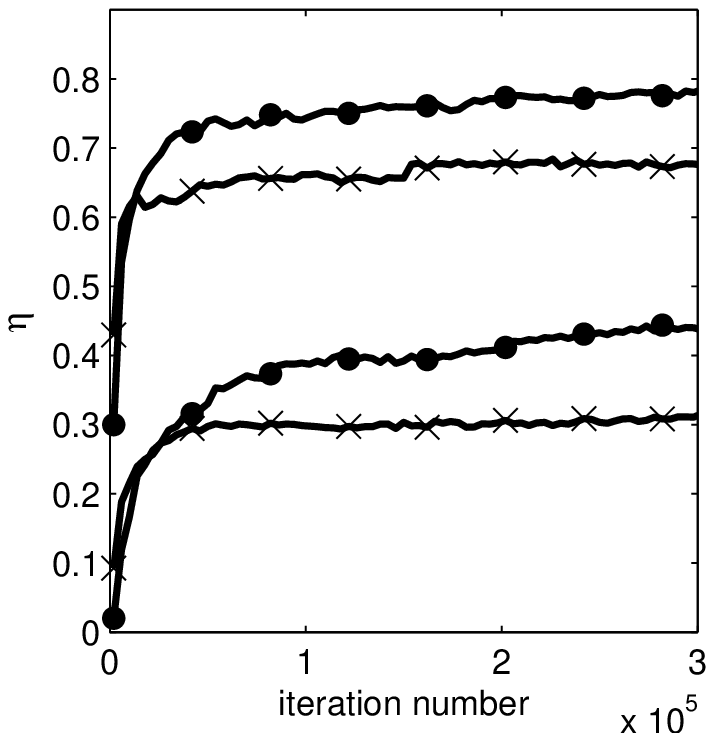}
\includegraphics[width=0.35\columnwidth]{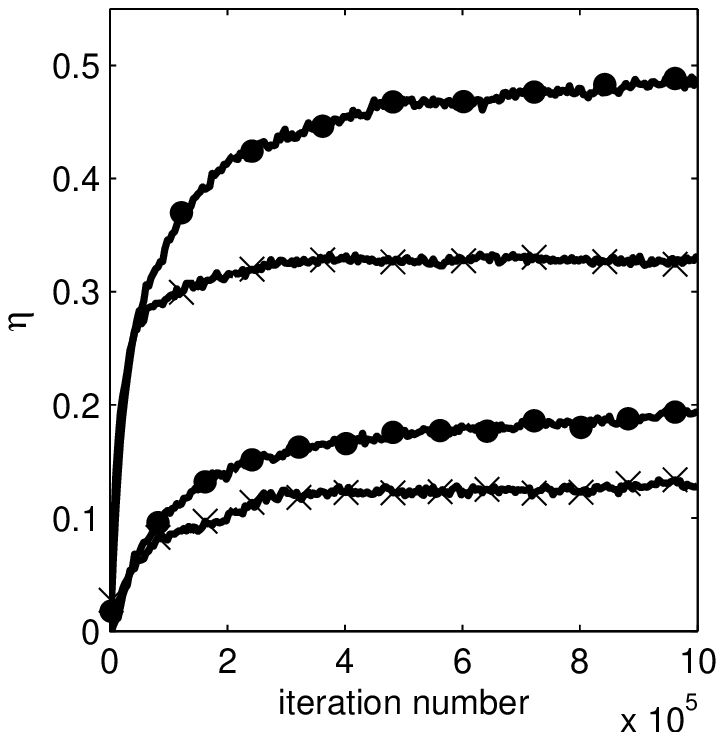}
}
\caption{Results for \textsc{garnet}$(30,4,2,0.1)$ (left pane) and \textsc{garnet}$(100,10,3,0.1)$ (right pane) where circled graphs are for adaptive bases. In each graph the lower two graphs are for $K_r=4$ and the upper graphs  are for $K_r=12$. See text for detail.}
\label{fig:1}
\end{center}
\vskip -0.5in
\end{figure}

\vspace{-0.1cm}
The \textsc{garnet}\footnote{brevity for Generic Average Reward Non-stationary Environment Test-bench} problems \cite{Archibald_etal1995,BhatnagarEtAl2009Report} are a class of randomly constructed finite MDPs serving as a test-bench for RL algorithms. A \textsc{garnet} problem is characterized by four parameters and is denoted by \textsc{garnet}$(X,U,B,\sigma)$. The parameter $X$ is the number of states, $U$ is the number of actions, $B$ is the branching factor, and $\sigma$ is the variance of each transition reward. When constructing such a problem, we generate for each state a reward, distributed according to $\mathcal{N}(0,1)$. For each state-action the reward is distributed according to $\mathcal{N}(g(x),\sigma^2)$. The transition matrix for each action is composed of
$B$ non-zero terms. We consider the same \textsc{garnet} problems as those simulated by \cite{BhatnagarEtAl2009Report}. For the critic's feature vector, we use the basis functions  $\phi(x,s) = \cos\left(\frac{x}{d}s + \varrho_{x,d}\right)$,
where  $x=1,\ldots, N$, $1\le d \le K_r$, $s\in \mathbb{R}^{1}$, and $\varrho_{x,d}$ are i.i.d. uniform random phases. Note that only one parameter in this simulation controls the basis functions. The actor's feature vectors are of size $K_a\times|U|$, and are constructed as
\begin{equation*}
\xi(x,u)  \triangleq(\overbrace{0,\ldots,0}^{K_a\times(u-1)}, \phi(x,s(t=0)),\overbrace{0,\ldots,0}^{K_a\times(|U|-u)}.
\end{equation*}
The policy function is $\mu(u|x,\theta) =e^{\theta^\top\xi(x,u)}/\sum_{u'\in U}e^{\theta^\top\xi(x,u')}$.
Bhatnagar et al. \cite{BhatnagarEtAl2009Report} reported simulation results for two \textsc{garnet} problems: \textsc{garnet}$(30,4,2,0.1)$ and \textsc{garnet}$(100,10,3,0.1)$. We based our simulations on these results where the time steps are identical to those of \cite{BhatnagarEtAl2009Report}. The \textsc{garnet}$(30,4,2,0.1)$ problem  (Fig. \ref{fig:1} left pane) was simulated for $K_r=4$ (two lower graphs) and $K_r=12$ (two upper graphs), where each graph is an average of $100$ repeats. The \textsc{garnet}$(100,10,3,0.1)$ problem  (Fig. \ref{fig:1} right pane) was simulated for $K_r=4$ (two lower graphs) and $K_r=12$ (two upper graphs), where each graph is an average of $100$ repeats. We can see that in such problems there is an evident advantage to an adaptive base, which can achieve additional fitness to the problem, and thus even for low dimensional problems the adaptation may be crucial.

\subsection{The Mountain Car}
\begin{figure}[ht!]
\begin{center}
\includegraphics[width=0.32\columnwidth]{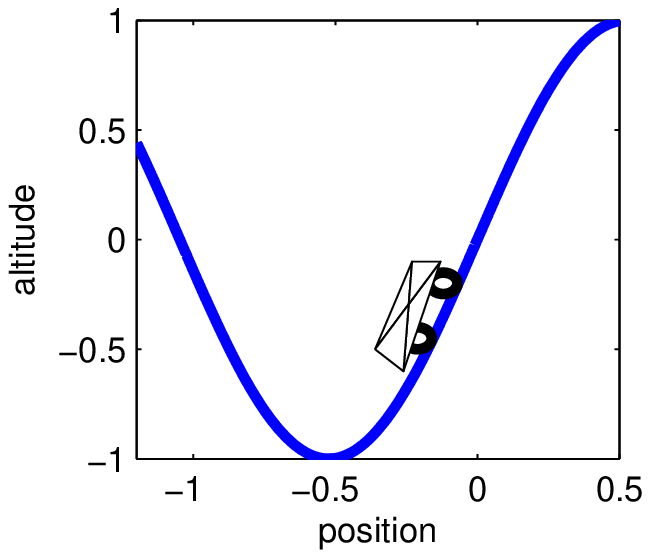}
\includegraphics[width=0.28\columnwidth]{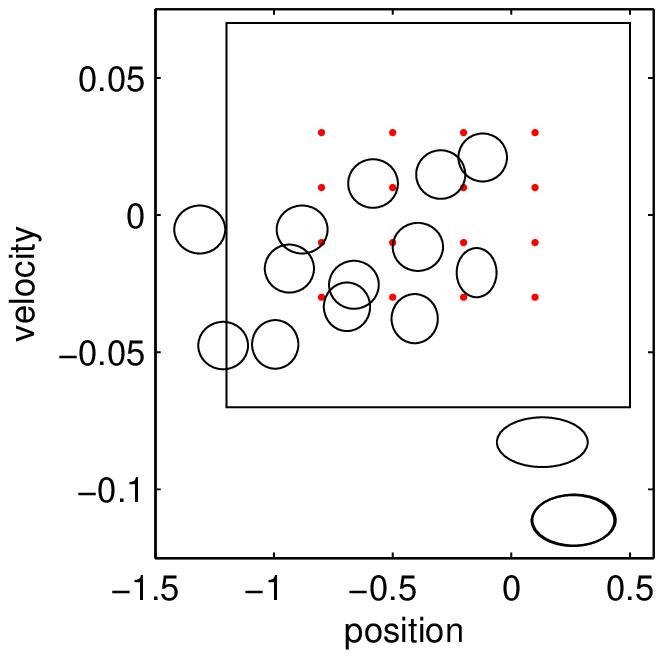}
\includegraphics[width=0.33\columnwidth]{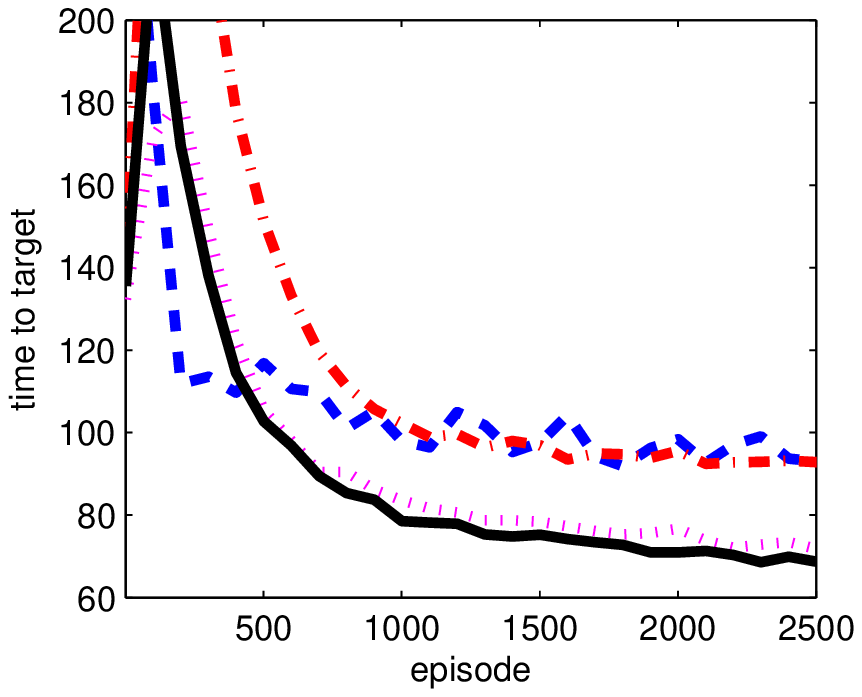}
\caption{\textbf{(left pane)} illustration of the mountain car task. \textbf{(middle pane)} Realization of ABTD with 16 basis functions where the red dots are the basis functions initial position and the circles are their final position. The radii are proportional to the variance. The rectangle represents the bounded parameter set of the car. \textbf{(right pane)} Simulation result for the mountain car problem with solutions of SARSA (blue dash) AC (red dash-dot) AB-AC with 64 basis functions (magenta dotted) AB-AC with 16 basis functions (black solid). }
\label{fig:2}
\end{center}
\vskip -0.2in
\end{figure}

The mountain car task (see \cite{SinghSutton1996} or \cite{SuttonBarto} for details) is a physical problem where a car is positioned randomly between two mountains (see Fig. \ref{fig:2} left pane) and needs to climb the right mountain, but the engine of the car does not support such a straight climb. Thus, the car needs to accumulate sufficient gradational energy, by applying back and forth actions, in order to succeed.

We applied the adaptive basis TD algorithm on this problem. We chose the critic basis functions to be radial basis functions (RBF) (see \cite{haykin2008}), where the value function is represented by $\sum_{i=1}^{M} r_i \exp\{-(p-s_i^{(p)})^2/s^2_{p,i} -(v-s_i^{(v)})^2/s^2_{v,i}\}$. The centers of the RBFs are parameterized by $(s_i^{(p)},s_i^{(v)})_{i=1}^M$ while the variance is represented by $(s^2_{p,i},s^2_{v,i})_{i=1}^{M}$. In the right pane of Fig. \ref{fig:2} we present simulation results for 4 cases: SARSA (blue dash) which is based on the implementation of \cite{SinghSutton1996}, AC (red dash-dot) with 64 basis functions uniformly distributed on the parameter space, ABTD with 64 basis functions (magenta dotted) where both the location and the variance of the basis functions can adapt, ABAC with 16 basis functions (black solid) with the same adaptation. We see that the adaptive basis gives a significant advantage in performance. Moreover, we see that even with small number of parameters, the performance is not affected. In the middle pane, the dynamics of a realization of the basis functions is presented where the dots and circles are the initial positions and final positions of the basis functions, respectively. The circle sizes are proportional to the basis functions standard deviations, i.e., $(s_{p,i},s_{v,i})_{i=1}^{M}$ .

\subsection{The Performance of Multiple Time Scales vs. Single Time Scale}
\begin{figure}[ht!]
\begin{center}
\includegraphics[width=0.4\columnwidth]{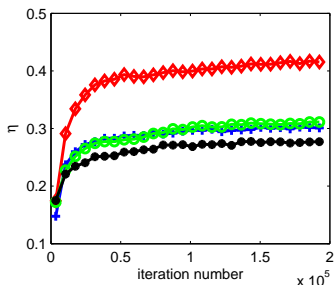}
\caption{Results for \textsc{garnet}$(30,5,5,0.1)$ for $K_r=8$. The upper diamond red graph is MTS ABTD algorithm, the circled green graph is STS ABTD acting on slow time scale, the blue crossed line is MTS static basis AC algorithm as in \cite{BhatnagarEtAl2009Report}, and the black stared line is STS ABTD acting on fast time scale. Each graph is average of 100 simulation runnings.
}
\label{fig:3}
\end{center}
\vskip -0.1in
\end{figure}

In this section we discuss the differences in performance between the MTS algorithm to the STS algorithms. Unlike mistakenly thought, neither MTS algorithms nor STS algorithms have advantage in terms of convergence. This difference comes from the fact that both methods perform the gradient algorithm differently, thus, they may result different trajectories. In Fig. \ref{fig:3} we can see a case on a \textsc{garnet}(30,5,5,0.1) where the MTS ABTD algorithm (upper red diamond graph) has an advantage over STS ABTD algorithms or MTS static basis AC algorithm as in \cite{BhatnagarEtAl2009Report} (rest of the graphs). We note that this is not always the case and it depends on the problem parameters or the initial conditions.
\section{Discussion} \label{sec:discussion}
We introduced three new AC based algorithms where the critic's basis is adaptive. Convergence proofs, in the average reward case, were provided. We note that the algorithms can be easily transformed to discounted reward. When considering other target functions, more AC algorithms with adaptive basis can be devised, e.g., considering the objective function $\|\mathrm{E}[d\phi]\|^2$ yields A$^\top$TD and GTD(0) algorithms \cite{SuttonEtAl2008NIPS}. Also, mixing the different algorithm introduced in here, can yield new algorithms with some desired properties. For example. we can devise an algorithm where the linear part is updated similar to \eqref{eq:TD_iterate_r} and the non-linear part is updated similar to \eqref{eq:TD_iterate_s_m}. Convergence of such algorithms will follow the same lines of proof as introduced here.

The advantage of adaptive bases is evident: they relieve the domain expert from the task of carefully designing the basis. Instead, he may choose a flexible basis, where one use algorithms as introduced here to adapt the basis to the problem at hand. From a methodological point of view, the method we introduced in this paper demonstrates how to easily transform an existing RL algorithm to an adaptive basis algorithm.
The analysis of the original problem is used to show convergence of the faster time scale and the slow time scale is used for modifying the basis, analogously to ``code reuse" concept in software engineering.


\end{document}